\documentclass[Afour,sageh,times]{sagej}

\usepackage{moreverb,url}
\usepackage{caption, subcaption}
\usepackage{makecell,rotating}
    \setcellgapes{5pt}
\usepackage{lipsum}
\usepackage{dblfloatfix}
\usepackage{multirow}

\usepackage[colorlinks,bookmarksopen,bookmarksnumbered,citecolor=red,urlcolor=red]{hyperref}

\newcommand\BibTeX{{\rmfamily B\kern-.05em \textsc{i\kern-.025em b}\kern-.08em
T\kern-.1667em\lower.7ex\hbox{E}\kern-.125emX}}

\def\volumeyear{2019}

\begin{document}

\runninghead{Woo-Ri Ko et al.}

\title{AIR-Act2Act: Human-human interaction dataset for teaching non-verbal social behaviors to robots}

\author{Woo-Ri Ko\affilnum{1},
Minsu Jang\affilnum{1},
Jaeyeon Lee\affilnum{1} and
Jaehong Kim\affilnum{1}}

\affiliation{\affilnum{1}Electronics and Telecommunications Research Institute (ETRI), KR}

\corrauth{Woo-Ri Ko, ETRI,
218 Gajeong-ro, Yuseong-gu, Daejeon, 34129, KR.}

\email{wrko@etri.re.kr}

\begin{abstract}
To better interact with users, a social robot should understand the users' behavior, infer the intention, and respond appropriately.
Machine learning is one way of implementing robot intelligence.
It provides the ability to automatically learn and improve from experience instead of explicitly telling the robot what to do.
Social skills can also be learned through watching human-human interaction videos.
However, human-human interaction datasets are relatively scarce to learn interactions that occur in various situations.
Moreover, we aim to use service robots in the elderly-care domain; however, there has been no interaction dataset collected for this domain.
For this reason, we introduce a human-human interaction dataset for teaching non-verbal social behaviors to robots.
It is the only interaction dataset that elderly people have participated in as performers.
We recruited 100 elderly people and two college students to perform 10 interactions in an indoor environment.
The entire dataset has 5,000 interaction samples, each of which contains depth maps, body indexes and 3D skeletal data that are captured with three Microsoft Kinect v2 cameras.
In addition, we provide the joint angles of a humanoid NAO robot which are converted from the human behavior that robots need to learn.
The dataset and useful python scripts are available for download at \href{https://github.com/ai4r/AIR-Act2Act}{https://github.com/ai4r/AIR-Act2Act}.
It can be used to not only teach social skills to robots but also benchmark action recognition algorithms.

\end{abstract}


\maketitle

\section{Introduction}
To better interact with users, a social robot should understand their behavior, infer the intention and formulate appropriate responses. 
For instance, if a user is crying in the bedroom, a robot should approach slowly and hug her shoulder gently.
Toward this need, many researchers, e.g. \cite{huang2012robot}, \cite{qureshi2016robot} and \cite{hemminahaus2017towards}, have focused on implementing social intelligence for robots.
However, these studies have a limitation that the robot repeats only predefined behaviors since they are more about behavior selection rather than behavior generation.

Machine learning is one way of overcoming the above limitation.
It provides the robots with the ability to automatically learn and improve from experience instead of explicitly telling them what to do.
In recent years, this methodology has shown good performances benefiting from the increased availability of demonstration data, as well as computational advancements brought on by deep learning.
\cite{wrko2019social} showed that it was feasible to a deep neural network for robots to learn social skills.
Two social behaviors, i.e., \textit{handshake} and \textit{wait}, were learned from 583 human-human interaction videos of \textit{NTU RGB+D} dataset introduced by \cite{shahroudy2016ntu}.
The dataset used was large enough to generate two behaviors, but a larger dataset is essential to generate more behaviors.
However, the existing datasets are not large enough to learn the interactions that occur in various situations.
Moreover, there has been no dataset collected for the elderly, so it is difficult to use in the elderly-care domain, which is what we are aiming to do.

For this reason, we introduce a human-human non-verbal interaction dataset, \textit{AIR-Act2Act}.
It was collected as a part of project \textit{AIR (AI for Robots)} which aims to provide socially assistive services to the elderly.
We recruited 100 elderly people and two college students to perform 10 interactions in an indoor environment.
With three Microsoft Kinect v2 cameras, depth maps, body indexes, and 3D skeletal data were captured concurrently.
In addition, the behaviors of the people to be learned were converted into the robot's joint angles.
In summary, the dataset has the following strengths:
\begin{enumerate}
    \item[(i)] It is the only interaction dataset of the elderly;
    \item[(ii)] It provides robotic data to be learned;
    \item[(iii)] It is one of the largest interaction datasets that provides 3D skeletal data;
    \item[(iv)] It can be used to not only teach social skills to robots but also benchmark action recognition algorithms.
\end{enumerate}

\begin{figure*}[ht]
    \centering
    \begin{subfigure}[b]{0.403\textwidth}
        \includegraphics[width=\textwidth]{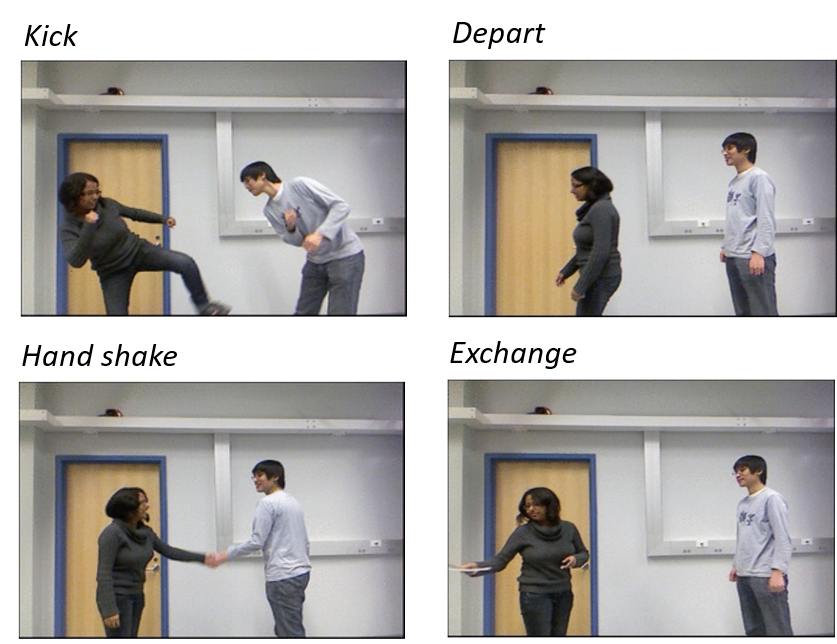}
        \caption{\centering SBU Kinect Interaction}
        \label{fig:sbu}
    \end{subfigure}
    \hfill%
    \begin{subfigure}[b]{0.47\textwidth}
        \includegraphics[width=\textwidth]{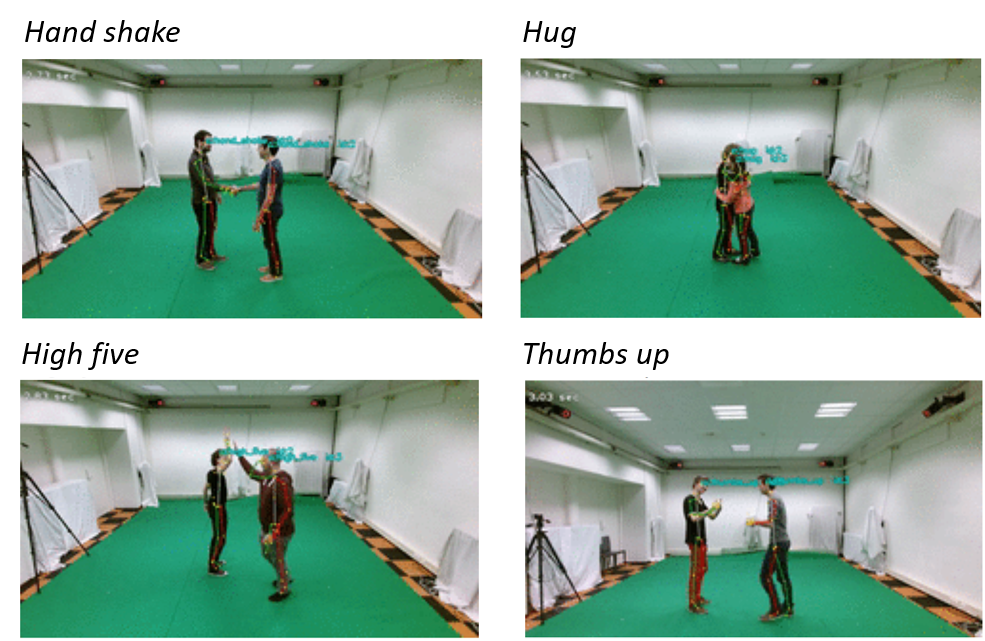}
        \caption{\centering ShakeFive2}
        \label{fig:shakefive2}
    \end{subfigure}
    \vskip\baselineskip
    \begin{subfigure}[b]{0.35\textwidth}
        \includegraphics[width=\textwidth]{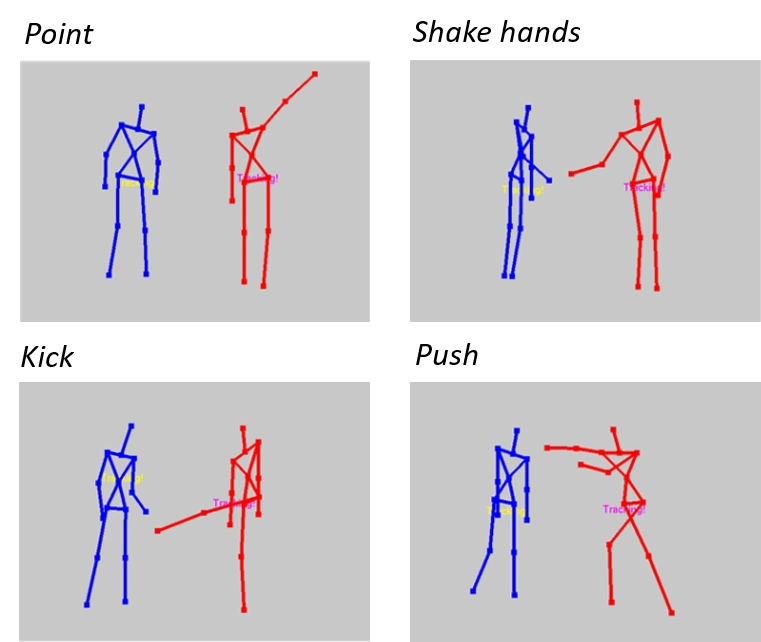}
        \caption{\centering K3HI}
        \label{fig:k3hi}
    \end{subfigure}
    \hfill%
    \begin{subfigure}[b]{0.35\textwidth}
        \includegraphics[width=\textwidth]{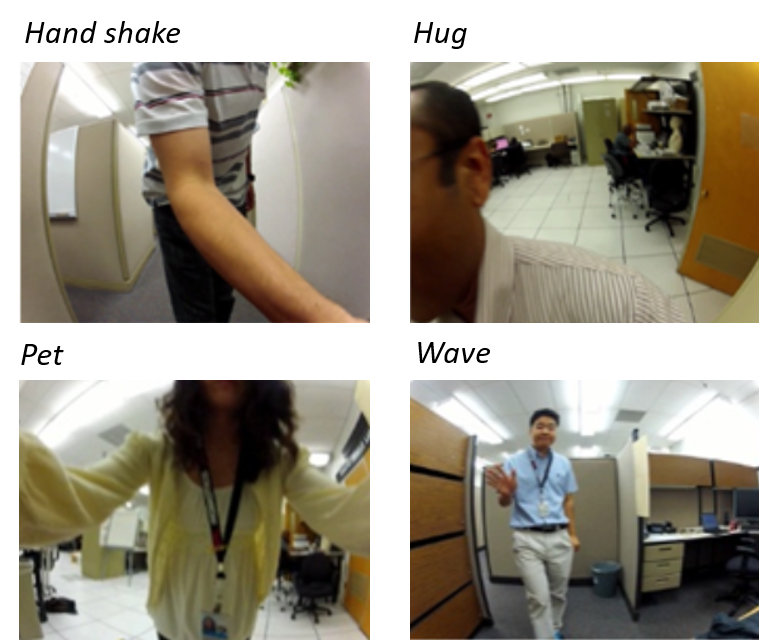}
        \caption{\centering JPL-Interaction}
        \label{fig:jpl}
    \end{subfigure}
    \hfill%
    \begin{subfigure}[b]{0.253\textwidth}
        \includegraphics[width=\textwidth]{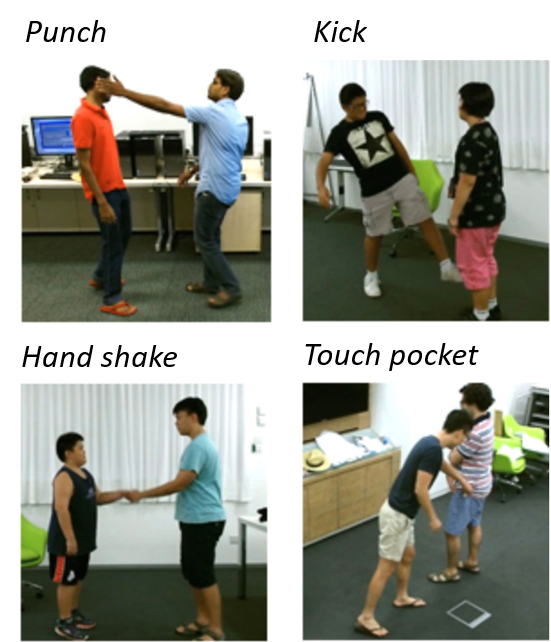}
        \caption{\centering NTU RGB+D 120}
        \label{fig:ntu}
    \end{subfigure}
    \caption{Related datasets provided for research purposes.}
    \label{fig:dataset}
\end{figure*}

\section{Related datasets}
Several human-human interaction datasets are accessible for research purposes, and they contain RGB videos, depth maps, and 3D skeletal data.
Example frames of the datasets appear in Figure \ref{fig:dataset}.

\subsection{SBU Kinect Interaction}
\cite{yun2012two} introduced the \textit{SBU Kinect Interaction} dataset captured by Microsoft Kinect.
It contains 300 videos of eight interactions: \textit{push, kick, punch, pass object, hug, hand shake, approach and depart.}
The RGB videos and depth maps were recorded in $640\times480$ resolution.
The skeletal data contain 3D coordinates of 15 joints per person.

\subsection{ShakeFive2}
\cite{van2016spatio} introduced a collection of human interaction clips captured by Microsoft Kinect v2 camera. 
There are eight interaction classes in the dataset: \textit{fist bump, handshake, high-five, hug, pass object, thumbs up, rock-paper-scissors and explain.}
Further, 153 videos were captured in lab conditions and encoded at a resolution of $1280\times720$.
The skeletal data contains 3D coordinates of 25 joints per person.

\subsection{K3HI}
\cite{hu2013efficient} introduced the \textit{K3HI} dataset captured by Kinect.
It provides 312 skeletal files of eight interactions: \textit{approach, depart, exchange, kick, point, punch, push, shake.}
The skeletal data contains 3D coordinates of 15 joints per person.

\subsection{JPL-Interaction} 
The \textit{JPL-Interaction} dataset is a first-person human-robot interaction dataset introduced by \cite{ryoo2015prediction}.
A GoPro2 camera is attached to the head of a humanoid model and human participants are asked to interact with the humanoid.
The dataset has eight videos containing 180 executions of seven activities: \textit{shake hand, hug, pet, wave hand, point, punch and throw object.}
The RGB videos and depth maps were recorded in $640\times480$ and $320\times240$ resolutions, respectively.
The skeletal data contains HOJ3D features obtained by \cite{xia2012view}.

\subsection{NTU RGB+D 120}
The \textit{NTU RGB+D 120} dataset is an action recognition dataset introduced by \cite{Liu_2019_NTURGBD120}.
It contains 8,276 action samples of 26 two-person actions, e.g. \textit{punch, kick, push, pat, etc.}
The RGB videos, depth maps and 3D skeletal data are captured by three Microsoft Kinect v2 concurrently.
The RGB videos and depth maps were recorded in $1920\times1080$ and $512\times424$ resolutions, respectively.
The skeletal data contains 3D coordinates of 25 major body joints per person.

\subsection{Other datasets without skeletal data}
From skeletal data, temporal patterns of interactions can be learned without having to consider viewpoint and person appearance.
However, some datasets contain RGB videos but no 3D skeletal data or depth maps depending on the cameras used.
If not available, the skeletal coordinates of the joints can be extracted from RGB frames by a 3D pose estimation algorithm such as that proposed by \cite{tome2017lifting}.
The datasets that can be used in this way are described as follows:

\begin{itemize}
    \item \textit{UT-Interaction}: \cite{ryoo2010ut} introduced videos of continuous executions of six classes of human-human interactions: \textit{shake-hands, point, hug, push, kick, and punch}.
    \item \textit{TV Human Interaction}: \cite{patron2010high} provided short video segments of four classes taken from popular TV series: \textit{handshake, hug, kiss, and highfive}.
    \item \textit{Hollywood2}: \cite{marszalek2009actions} introduced movie clips of four classes: \textit{fight, handshake, hug, and kiss}.
    \item \textit{DeepMind Kinetics}: \cite{kay2017kinetics} provided 10-s clips from YouTube videos of 11 interaction classes including \textit{handshake, hug, and massage feet}.
\end{itemize}

\section{Our dataset: AIR-Act2Act}
In this section, we introduce the details of the \textit{AIR-Act2Act} dataset, which is the only human-human interaction dataset that elderly people have participated in as performers.

\subsection{Subjects}
We recruited 100 elderly people and two college students for our data collection.
The elderly people were recruited at the senior welfare centers, based
on the following criteria: (1) age over 60; (2) healthy enough to stand and walk for a few minutes.
The mean age of the elderly people was 77 years (ranging from 64 to 88) and 39$\%$ were males.
The college students were recruited online to perform as partners of the elderly.
All subjects completed an institutional review board (IRB)-approved consent form prior to participating in a data acquisition session.

\subsection{Scenarios}
We asked participants to perform each scenario, described in Table \ref{tab:class}, five times.
Each interaction scenario is defined as a pair of coordinated behaviors: an \textit{initiating} behavior performed by an elderly person, and a \textit{responsive} behavior performed by a partner.
The initiating behaviors consisted of eight greeting behaviors motivated by \cite{heenan2014designing} and an additional two behaviors of \textit{high-five} and \textit{hit}.
The responsive behaviors were designed so that, when performed by service robots, they would be acceptable to people as natural and humble reactions.
Since we did not instruct the participants to act in an exact pattern, there were large variations in intra-class action trajectories.

\begin{table}[ht]
    \small\sf\centering
    \caption{10 interaction scenarios. (E: elderly person, R: partner performing as a robot) \label{tab:class}}
    \begin{tabular}{cl}
    \toprule
    & Scenario \\ \midrule
    \multirow{2}{*}{1} & E: enters into the service area through the door. \\
    & R: bows to the elderly person. \\ \midrule
    \multirow{2}{*}{2} & E: stands still without a purpose. \\
    & R: stares at the elderly person for a command. \\ \midrule
    \multirow{2}{*}{3} & E: calls the robot. \\
    & R: approaches the elderly person. \\ \midrule
    \multirow{2}{*}{4} & E: stares at the robot. \\
    & R: scratches its head from awkwardness. \\ \midrule
    \multirow{2}{*}{5} & E: lifts his arm to shake hands. \\
    & R: shakes hands with the elderly person. \\ \midrule
    \multirow{2}{*}{6} & E: covers his face and cries. \\
    & R: stretches his hands to hug the elderly person. \\ \midrule
    \multirow{2}{*}{7} & E: lifts his arm for a high-five. \\
    & R: high-fives with the elderly person. \\ \midrule
    \multirow{2}{*}{8} & E: threatens to hit the robot. \\
    & R: blocks the face with arms. \\ \midrule
    \multirow{2}{*}{9} & E: beckons to go away. \\
    & R: turns back and leaves the service area. \\ \midrule
    \multirow{2}{*}{10} & E: turns back and walks to the door. \\
    & R: bows to the elderly person. \\
    \bottomrule
    \end{tabular}
\end{table}

\subsection{Collection setups}
\begin{figure}[ht]
    \centering
    \begin{subfigure}{\columnwidth}
        \includegraphics[height=7cm]{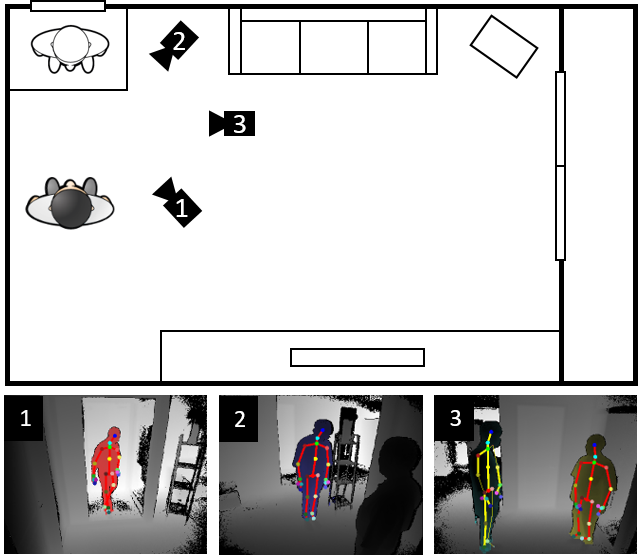} 
        \caption{\centering Scenarios 1 and 10}
        \label{fig:daejeon1}
    \end{subfigure}
    \par\bigskip
    \begin{subfigure}{\columnwidth}
        \includegraphics[height=7cm]{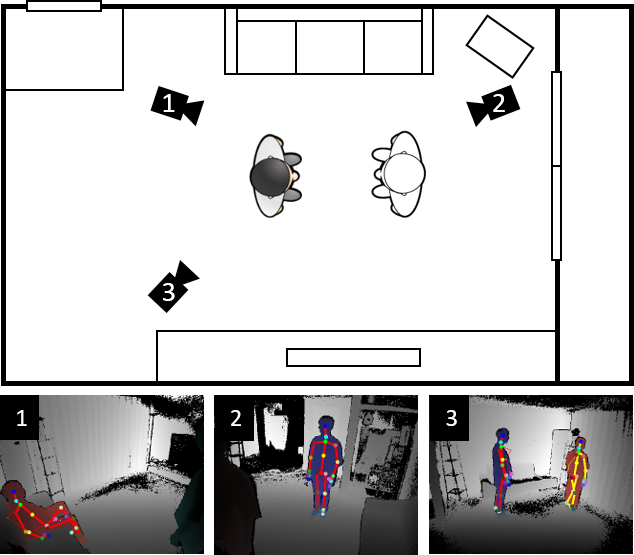} 
        \caption{\centering Scenarios 2, 3 and 4}
        \label{fig:daejeon2}
    \end{subfigure}
    \par\bigskip
    \begin{subfigure}{\columnwidth}
        \includegraphics[height=7cm]{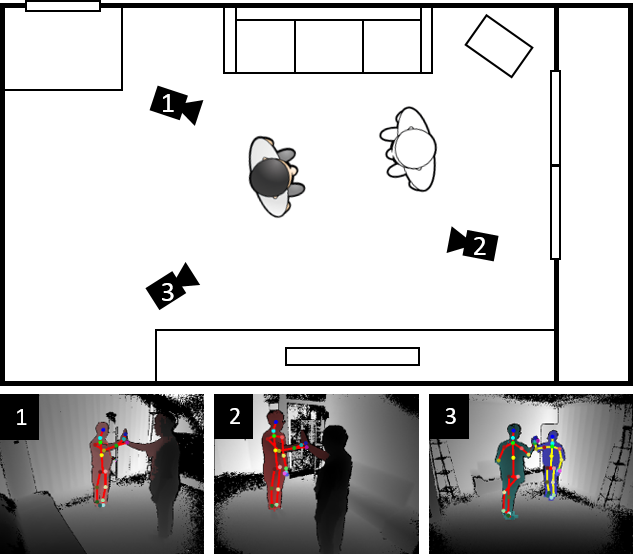} 
        \caption{\centering Scenarios 5, 6, 7, 8 and 9}
        \label{fig:daejeon3}
    \end{subfigure}
    \setlength{\belowcaptionskip}{-10pt}
    \caption{The locations of participants and cameras in the apartment environment. The elderly person and his/her partner are denoted as white and colored people, respectively.}
    \label{fig:daejeon}
\end{figure}

\begin{figure}[ht]
    \centering
    \includegraphics[height=7cm]{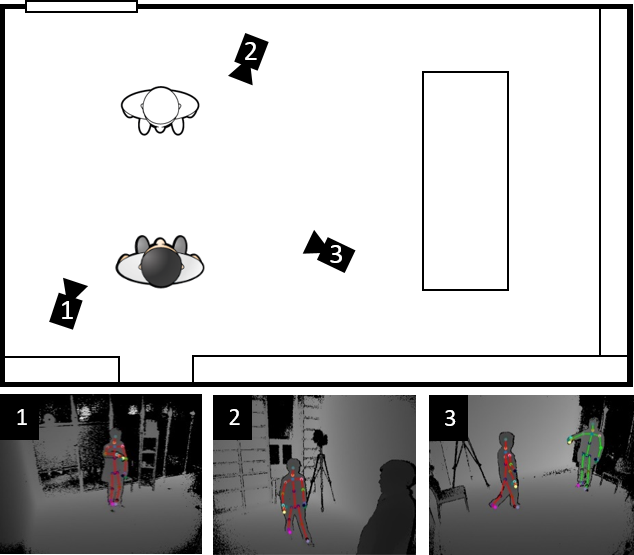} 
    \caption{The senior welfare center environment.}
    \label{fig:gunpo}
\end{figure}

Our interaction data were collected in an apartment and a senior welfare center where service robots are likely to be used.
Figure \ref{fig:daejeon} shows the locations of participants and cameras in the apartment environment.
Scenarios 1 and 10 were performed at the front door to allow elderly people to enter and exit through the door, and the other scenarios were performed in the living room, which was equipped with a TV and a sofa.
Figure \ref{fig:gunpo} shows the senior welfare center environment, which is a meeting room with a door.
For each scenario, three cameras were set up at the same height; however, were positioned to capture different views.
Two cameras were placed next to each person to capture the behaviors from the other person's point of view.
The last camera was placed in a position where both participants were visible in order to gather information of the participants relative to each other.
The position of each camera was adjusted each time to take into consideration the movement range of the parrticipants.
In total, the entire dataset has 5,000 interaction samples with three different views, where each view lasts for about 6 s.

\subsection{Data modalities}
We used three Microsoft Kinect v2 sensors to collect our dataset.
Each Kinect sensor provided depth maps, body indexes, and 3D skeletal data by using the APIs of \cite{kinect}.
The depth maps and body indexes have a resolution of $512\times424$ pixels and a frame rate of 30 fps.
The 3D coordinates of 25 joints were obtained via trained and randomized decision tree forests proposed in \cite{shotton2011real}.
We refined the skeletal data to check the existing tracking failures by manually checking the whole dataset.
In addition, the human behaviors to be learned are transformed into the behaviors of a NAO robot.
The detailed information of each data file is explained in the following.

\subsection{Description of depth map}
The depth map is a sequence of two-dimensional depth values in millimeters, where each of the depth value ranges from 0 to 8,000 mm.
Each individual map is stored in a separate \textit{.png} file, which is a 16-bit grayscale image with a resolution of $512\times424$ pixels.
Figures \ref{fig:daejeon} and \ref{fig:gunpo} show an example of depth maps displayed in grayscale.

\subsection{Description of body indexes}
The body index frame indicates which of the tracked subjects the depth pixels belong to.
Since Kinect sensors can detect up to six people, the body index of each person is set to a value between 0 and 5; any other value indicates that it is the background.
Each individual body index frame is stored in a separate \textit{.png} file which is an 8-bit grayscale image.
Figure \ref{fig:daejeon} shows an example of body indexes denoted by different colors on the depth maps.
Note that the body indexes were captured only in the apartment environment.

\begin{table*}[!b]
\small\sf\centering
\caption{Comparison of our dataset with existing human-human interaction datasets. (D: depth, S: skeleton, B: body index)}
\begin{tabular}{lrrrclcl}
\toprule
& $\#$Subjects & $\#$Actions & $\#$Samples & $\#$Views & Data Modalities & Year & Description \\ 
\midrule
\textit{Hollywood2} & - & \textbf{12} & 2,517 & 1 & RGB & 2009 & YouTube \\ 
\textit{UT-Interaction} & 6 & 6 & 120 & 1 & RGB & 2010 & Outdoor\\
\textit{TV Human Interaction} & - & 4 & 300 & 1 & RGB & 2010 & YouTube\\ 
\textit{SBU Kinect Interaction} & 7 & 8 & 300 & 1 & RGB+D+S & 2012 & \\
\textit{K3HI} & 15 & 8 & 312 & 1 & S & 2013 & \\
\textit{JPL-Interaction} & 8 & 9 & 180 & 1 & RGB+S & 2015 & Human-robot \\
\textit{ShakeFive2} & 33 & 8 & 153 & 1 & RGB+S & 2016 & \\
\textit{DeepMind Kinetics} & - & \textbf{11} & \textbf{6,378} & 1 & RGB & 2017 & YouTube \\ 
\textit{NTU RGB+D 120} & \textbf{106} & \textbf{26} & \textbf{8,276} & \textbf{3} & RGB+D+S & 2019 & \\
\midrule
\textit{\textbf{AIR-Act2Act}} & \textbf{100} & \textbf{10} & \textbf{5,000} & \textbf{3} & D+S+B+Robotic & 2019 & The elderly \\  
\bottomrule
\end{tabular}
\label{tab:dataset}
\end{table*}

\subsection{Description of skeletal data}
The skeletal information consists of 3D coordinates of 25 major body joints for each detected human body in the scene.
The original skeletal data were stored in a \textit{.$\sim$joint} file (JSON format) and contained the following information for each body:
\begin{itemize}
    \item $bodyID$: tracking ID for the body;
    \item $trackingState$: tracking state of the body;
    \item $leanX$, $leanY$: lean vector of the body;
    \item $x_j$, $y_j$, $z_j$: 3D location of the $j$th joint in camera space;
    \item $trackingState_j$: tracking state of the $j$th joint;
    \item $orientationX_j$, $orientationY_j$, $orientationZ_j$, \\ $orientationW_j$: orientation of the $j$th joint;
    \item $depthX_j$, $depthY_j$: 2D location of the $j$th joint on the depth map.
    \item $depthZ_j$: depth value of the $j$th joint.
\end{itemize}
The example skeletons are denoted by dots and lines on the depth maps in Figures \ref{fig:daejeon} and \ref{fig:gunpo}.

\subsection{Refinement of skeletal data}
\begin{figure}[ht]
    \centering
    \includegraphics[width=\columnwidth]{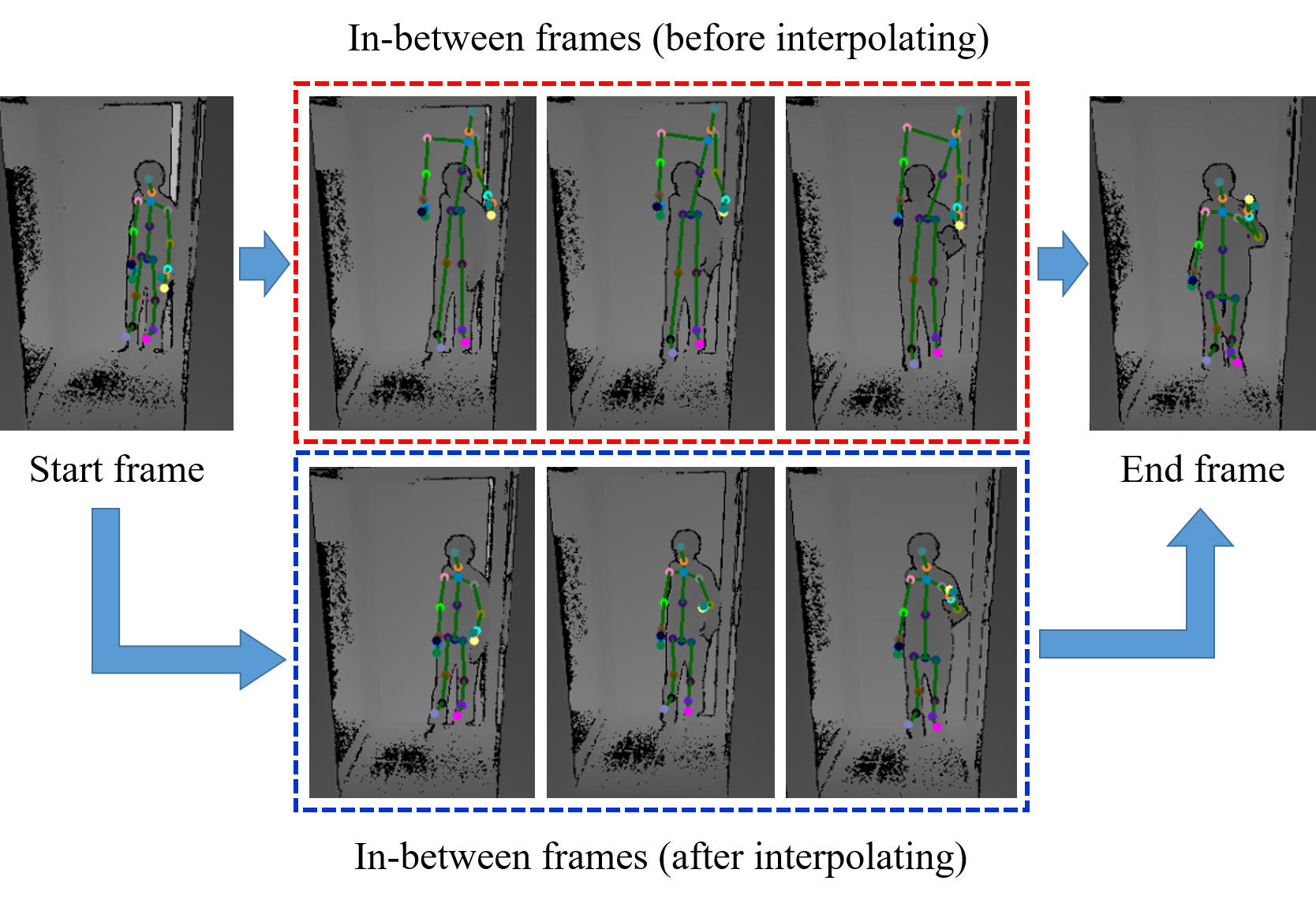}
    \caption{Example frames before and after the interpolation.}
    \label{fig:interpolate}
\end{figure}

The skeletons extracted by the Kinect sensor are mostly accurate; however, there are some tracking failures in some frames.
To account for this, we manually refined the skeletal data and stored it in a \textit{.joint} file (JSON format).
Firstly, we restored the incorrectly inferred or missing skeleton by interpolating the front and back frames.
Figure \ref{fig:interpolate} shows the example frames before and after the interpolation.

\begin{figure}[ht]
    \begin{subfigure}{.3\columnwidth}
    \centering
        \includegraphics[width=\columnwidth]{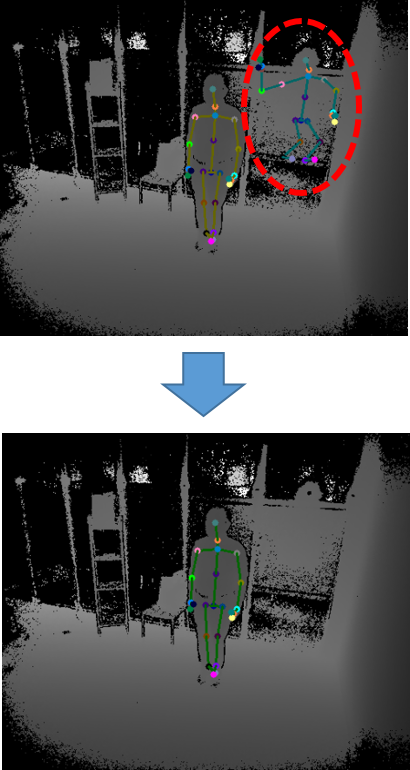} 
        \caption{\centering Camera 1}
    \end{subfigure}
    \hspace{0.2cm}
    \begin{subfigure}{.3\columnwidth}
    \centering
        \includegraphics[width=\columnwidth]{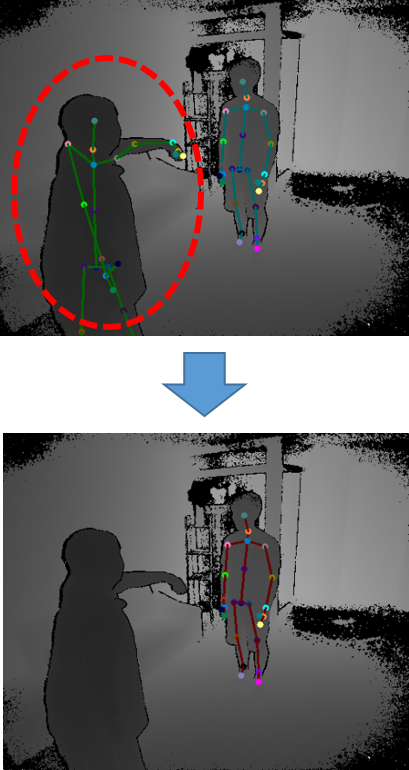} 
        \caption{\centering Camera 2}
    \end{subfigure}
    \hspace{0.2cm}
    \begin{subfigure}{.3\columnwidth}
    \centering
        \includegraphics[width=\columnwidth]{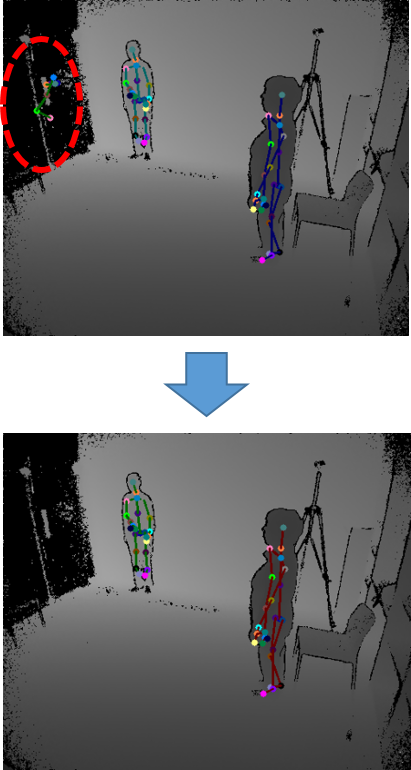} 
        \caption{\centering Camera 3}
    \end{subfigure}
    \caption{Example frames before and after removing skeletons.}
    \label{fig:remove}
\end{figure}

Secondly, additional skeletons other than those the camera should capture were removed, as shown in Figure \ref{fig:remove}.
This was done to ensure that in the videos taken by camera 1, there was only the skeleton of the elderly participant remaining, and in the videos taken by camera 2, there was only the skeleton of the partner remaining. 
In the videos taken by camera 3, the additional skeleton was removed to retain the skeletons of the elderly participant and their partner only.

Finally, if the 3D joint locations in the camera space, i.e., $x_j$, $y_j$ and $z_j$ (floats), were missing or incorrectly inferred, they were re-inferred from the 2D joint locations in the depth space and depth values.
The 2D locations and depth values were manually refined in advance.

\subsection{Generation of robotic data}
\begin{figure}[ht]
    \begin{subfigure}{.49\columnwidth}
    \centering
        \includegraphics[height=4cm]{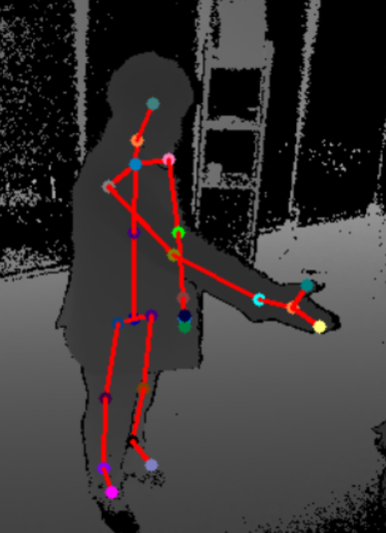}
        \caption{\centering Partner of an elderly person}
    \end{subfigure}
    \begin{subfigure}{.49\columnwidth}
    \centering
        \includegraphics[height=4cm]{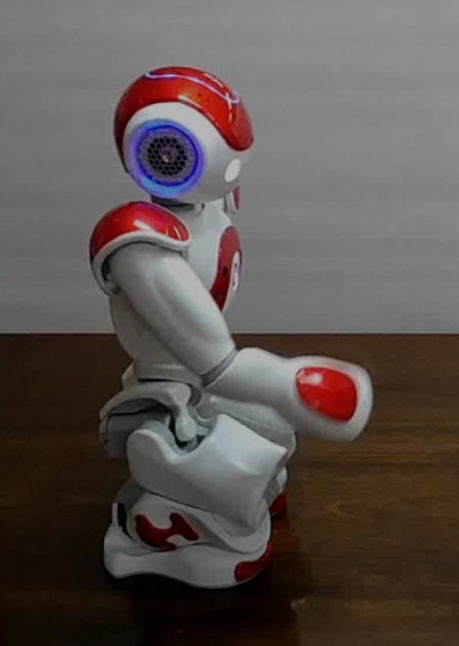} 
        \caption{\centering NAO robot}
    \end{subfigure}
    \caption{Example 3D skeletal data of a partner, and the transformed pose of a NAO robot.}
    \label{fig:robotic}
\end{figure}

Since the purpose of our dataset is to train robot intelligence, we transformed the behavior of the human that robots should learn, i.e., the partner of the elderly person, into the action of the robot.
We choose the NAO robot as the target robot platform.
This is a humanoid robot developed by \cite{NAO}.
It is designed to show human-like movements with two arms, two legs and a head.
Because of the balancing problem, we did not consider the lower body movement; therefore, 10 joint angles, i.e., pitch of hip and head, pitch and roll of L/R shoulders, yaw and roll of L/R elbows, were analytically calculated.
The joint angles were stored in a \textit{.nao} file (JSON format).
Figure \ref{fig:robotic} shows an example of the 3D skeletal data of a partner and the transformed pose of a NAO robot.

\subsection{Comparison with other datasets}
The comparison of our dataset with the existing human-human interaction datasets is presented in Table \ref{tab:dataset}.
The numbers of subjects, actions, samples and views, and data modalities are summarized.
Our dataset is the only interaction dataset of the elderly that provides robotic data to be learned.
Moreover, it is one of the largest datasets, containing 5,000 interaction samples of 100 subjects.
Each interaction sample provides depth maps, body indexes, and 3D skeletal data captured from three different points of view.

\section{Data usage}
Our dataset can be used to train various robot intelligence.

\subsection{Social behavior generation}
The dataset provides the skeletal data of two human participants that interact with each other.
Among them, the poses of an elderly person can be used as training input to determine which behavior to generate.
At the same time, the poses of the other person can be used as the ground truth robot behaviors after retargeting to a humanoid robot.

\subsection{Human action recognition}
The dataset provides the videos captured from three different points of view.
One is the third-person's point of view and the other two are the first-person point of view of each participant.
The videos captured from the third person's point of view can be used as training input to recognize interactive behaviors, and the rest to recognize one-person actions.
At the same time, the behaviors defined in Table \ref{tab:class} can be used as recognition results.

\section{Data access method}
Each subset of the \textit{AIR-Act2Act} and useful python scripts are available for download at \href{https://github.com/ai4r/AIR-Act2Act}{https://github.com/ai4r/AIR-Act2Act}.
The scripts include loading data into arrays and plotting skeletal data on depth maps, which will serve as good examples of how to utilize the dataset.
The information of the license details can also be found on the webpage.

\begin{acks}
This work was supported by the Institute of Information $\&$ communications Technology Planning $\&$ Evaluation (IITP) grant funded by the Korea government (MSIT) (No. 2017-0-00162, Development of Human-care Robot Technology for Aging Society)
\end{acks}

\bibliographystyle{SageH}
\bibliography{reference.bib}

\begin{thebibliography}{20}
\providecommand{\natexlab}[1]{#1}
\providecommand{\url}[1]{\texttt{#1}}
\providecommand{\urlprefix}{URL }
\expandafter\ifx\csname urlstyle\endcsname\relax
  \providecommand{\doi}[1]{DOI:\discretionary{}{}{}#1}\else
  \providecommand{\doi}{DOI:\discretionary{}{}{}\begingroup
  \urlstyle{rm}\Url}\fi

\bibitem[{{Aldebaran Robotics}(2012)}]{NAO}
{Aldebaran Robotics} (2012) Nao.
\newblock \url{http://doc.aldebaran.com/2-1/home_nao.html}.

\bibitem[{Heenan et~al.(2014)Heenan, Greenberg, Aghel-Manesh and
  Sharlin}]{heenan2014designing}
Heenan B, Greenberg S, Aghel-Manesh S and Sharlin E (2014) Designing social
  greetings in human robot interaction.
\newblock In: \emph{Proceedings of the 2014 conference on Designing interactive
  systems}. ACM, pp. 855--864.

\bibitem[{Hemminahaus and Kopp(2017)}]{hemminahaus2017towards}
Hemminahaus J and Kopp S (2017) Towards adaptive social behavior generation for
  assistive robots using reinforcement learning.
\newblock In: \emph{2017 12th ACM/IEEE International Conference on Human-Robot
  Interaction (HRI}. IEEE, pp. 332--340.

\bibitem[{Hu et~al.(2013)Hu, Zhu, Guo and Su}]{hu2013efficient}
Hu T, Zhu X, Guo W and Su K (2013) Efficient interaction recognition through
  positive action representation.
\newblock \emph{Mathematical Problems in Engineering} 2013.

\bibitem[{Huang and Mutlu(2012)}]{huang2012robot}
Huang CM and Mutlu B (2012) Robot behavior toolkit: generating effective social
  behaviors for robots.
\newblock In: \emph{2012 7th ACM/IEEE International Conference on Human-Robot
  Interaction (HRI)}. IEEE, pp. 25--32.

\bibitem[{Kay et~al.(2017)Kay, Carreira, Simonyan, Zhang, Hillier,
  Vijayanarasimhan, Viola, Green, Back, Natsev et~al.}]{kay2017kinetics}
Kay W, Carreira J, Simonyan K, Zhang B, Hillier C, Vijayanarasimhan S, Viola F,
  Green T, Back T, Natsev P et~al. (2017) The kinetics human action video
  dataset.
\newblock \emph{arXiv preprint arXiv:1705.06950} .

\bibitem[{Ko et~al.(2018)Ko, Yoon, Jang, Lee and Kim}]{wrko2019social}
Ko WR, Yoon Y, Jang M, Lee J and Kim J (2018) End-to-end learning-based
  interaction behavior generation for social robots.
\newblock In: \emph{ICSR2018 Workshop on Social Human-Robot Interaction of
  Service Robots}.

\bibitem[{Liu et~al.(2019)Liu, Shahroudy, Perez, Wang, Duan and
  Kot}]{Liu_2019_NTURGBD120}
Liu J, Shahroudy A, Perez M, Wang G, Duan LY and Kot AC (2019) Ntu rgb+d 120: A
  large-scale benchmark for 3d human activity understanding.
\newblock \emph{IEEE Transactions on Pattern Analysis and Machine Intelligence}
  \doi{10.1109/TPAMI.2019.2916873}.

\bibitem[{Marszalek et~al.(2009)Marszalek, Laptev and
  Schmid}]{marszalek2009actions}
Marszalek M, Laptev I and Schmid C (2009) Actions in context.
\newblock In: \emph{Computer Vision and Pattern Recognition, 2009. CVPR 2009.
  IEEE Conference on}. IEEE, pp. 2929--2936.

\bibitem[{{Microsoft Corp.}(2014)}]{kinect}
{Microsoft Corp} (2014) {Kinect for Windows SDK 2.0 Documentation}.

\bibitem[{Patron-Perez et~al.(2010)Patron-Perez, Marszalek, Zisserman and
  Reid}]{patron2010high}
Patron-Perez A, Marszalek M, Zisserman A and Reid ID (2010) {High Five:
  Recognising human interactions in TV shows.}
\newblock In: \emph{BMVC}, volume~1. Citeseer, p.~2.

\bibitem[{Qureshi et~al.(2016)Qureshi, Nakamura, Yoshikawa and
  Ishiguro}]{qureshi2016robot}
Qureshi AH, Nakamura Y, Yoshikawa Y and Ishiguro H (2016) Robot gains social
  intelligence through multimodal deep reinforcement learning.
\newblock In: \emph{2016 IEEE-RAS 16th International Conference on Humanoid
  Robots (Humanoids)}. IEEE, pp. 745--751.

\bibitem[{Ryoo and Aggarwal(2010)}]{ryoo2010ut}
Ryoo MS and Aggarwal J (2010) {UT-interaction dataset, ICPR contest on semantic
  description of human activities (SDHA)}.
\newblock In: \emph{IEEE International Conference on Pattern Recognition
  Workshops}, volume~2. p.~4.

\bibitem[{Ryoo et~al.(2015)Ryoo, Fuchs, Xia, Aggarwal and
  Matthies}]{ryoo2015prediction}
Ryoo MS, Fuchs TJ, Xia L, Aggarwal JK and Matthies L (2015) Robot-centric
  activity prediction from first-person videos: What will they do to me?
\newblock In: \emph{ACM/IEEE International Conference on Human-Robot
  Interaction (HRI)}. Portland, OR, pp. 295--302.

\bibitem[{Shahroudy et~al.(2016)Shahroudy, Liu, Ng and Wang}]{shahroudy2016ntu}
Shahroudy A, Liu J, Ng TT and Wang G (2016) {NTU RGB+D: A large scale dataset
  for 3D human activity analysis}.
\newblock In: \emph{Proceedings of the IEEE conference on computer vision and
  pattern recognition}. pp. 1010--1019.

\bibitem[{Shotton et~al.(2011)Shotton, Fitzgibbon, Cook, Sharp, Finocchio,
  Moore, Kipman and Blake}]{shotton2011real}
Shotton J, Fitzgibbon A, Cook M, Sharp T, Finocchio M, Moore R, Kipman A and
  Blake A (2011) Real-time human pose recognition in parts from single depth
  images.
\newblock In: \emph{Computer Vision and Pattern Recognition (CVPR), 2011 IEEE
  Conference on}. IEEE, pp. 1297--1304.

\bibitem[{Tome et~al.(2017)Tome, Russell and Agapito}]{tome2017lifting}
Tome D, Russell C and Agapito L (2017) Lifting from the deep: Convolutional 3d
  pose estimation from a single image.
\newblock In: \emph{Proceedings of the IEEE Conference on Computer Vision and
  Pattern Recognition}. pp. 2500--2509.

\bibitem[{Van~Gemeren et~al.(2016)Van~Gemeren, Poppe and
  Veltkamp}]{van2016spatio}
Van~Gemeren C, Poppe R and Veltkamp RC (2016) Spatio-temporal detection of
  fine-grained dyadic human interactions.
\newblock In: \emph{International Workshop on Human Behavior Understanding}.
  Springer, pp. 116--133.

\bibitem[{Xia et~al.(2012)Xia, Chen and Aggarwal}]{xia2012view}
Xia L, Chen CC and Aggarwal JK (2012) View invariant human action recognition
  using histograms of 3d joints.
\newblock In: \emph{2012 IEEE Computer Society Conference on Computer Vision
  and Pattern Recognition Workshops}. IEEE, pp. 20--27.

\bibitem[{Yun et~al.(2012)Yun, Honorio, Chattopadhyay, Berg and
  Samaras}]{yun2012two}
Yun K, Honorio J, Chattopadhyay D, Berg TL and Samaras D (2012) Two-person
  interaction detection using body-pose features and multiple instance
  learning.
\newblock In: \emph{Computer Vision and Pattern Recognition Workshops (CVPRW),
  2012 IEEE Computer Society Conference on}. IEEE, pp. 28--35.

\end{thebibliography}

\end{document}